\documentclass[runningheads]{llncs}
\usepackage[T1]{fontenc}
\usepackage{graphicx}
\usepackage{amsmath}
\usepackage{amsfonts}
\usepackage{multirow}
\usepackage{hyperref}
\usepackage{caption}
\usepackage{subcaption}
\DeclareMathOperator*{\argmin}{arg\,min}

\usepackage{xcolor}
\definecolor{color1}{HTML}{000000} 
\definecolor{color2}{HTML}{004949} 
\definecolor{color3}{HTML}{009292} 
\definecolor{color4}{HTML}{ff6db6} 
\definecolor{color5}{HTML}{ffb6db} 
\definecolor{color6}{HTML}{490092} 
\definecolor{color7}{HTML}{006ddb} 
\definecolor{color8}{HTML}{b66dff} 
\definecolor{color9}{HTML}{6db6ff} 
\definecolor{color10}{HTML}{b6dbff} 
\definecolor{color11}{HTML}{920000} 
\definecolor{color12}{HTML}{924900} 
\definecolor{color13}{HTML}{db6d00} 
\definecolor{color14}{HTML}{24ff24} 
\definecolor{color15}{HTML}{ffff6d}  

\usepackage{tikz}
\usetikzlibrary{arrows}
\usetikzlibrary{patterns}
\usetikzlibrary{positioning}
\usetikzlibrary{shapes.geometric}
\usetikzlibrary{shapes.misc}
\usetikzlibrary{calc}
\usetikzlibrary{arrows}
\usetikzlibrary{patterns}
\usetikzlibrary{positioning}
\usetikzlibrary{backgrounds, calc,fit}
\usetikzlibrary{shapes.misc}

\def\normaltwo{\x,{exp((-(\x)^2)/0.5)}}
\newsavebox{\genericfilt}
\savebox{\genericfilt}{%
\begin{tikzpicture}[font=\small,>=stealth,yscale=0.11,xscale=0.15,rotate=270]
  \draw[line width=0.15mm,domain=-1.5:1.5] plot (\normaltwo);
\end{tikzpicture}%
}

\def\G{\mathcal{G}}
\newcommand{\Node}{\mathrm{N}} 
\newcommand{\Sum}{\mathrm{S}} 
\newcommand{\Prod}{\mathrm{P}} 
\DeclareMathOperator{\ch}{ch}

\def\vz{\mathbf{z}}

\newlength{\figsize}
\def\twidth{1mm}

\makeatletter
\newcommand{\printfnsymbol}[1]{%
  \noindent\textsuperscript{\@fnsymbol{#1}}%
}
\makeatother

\begin{document}
\title{Anomaly Detection using Generative Models and Sum-Product Networks in Mammography Scans}
\titlerunning{Mammography Anomaly Detection by Generative Models and SPNs}
\author{Marc Dietrichstein \inst{1}\thanks{Equal contribution} \and
David Major \inst{1}\printfnsymbol{1} \and
Martin Trapp \inst{2} \and
Maria Wimmer \inst{1} \and
Dimitrios Lenis \inst{1} \and
Philip Winter \inst{1} \and
Astrid Berg \inst{1} \and
Theresa Neubauer \inst{1} \and
Katja~B\"uhler \inst{1}}
\authorrunning{M. Dietrichstein and D. Major et al.}
\institute{VRVis Zentrum f\"ur Virtual Reality und Visualisierung Forschungs-GmbH, \\ Vienna, Austria \\ 
\email{david.major@vrvis.at} \and
Department of Computer Science, Aalto University, Espoo, Finland
\thanks{This preprint has not undergone peer review (when applicable) or any post-submission improvements or corrections. The Version of Record of this contribution is published in LNCS 13609, and is available online at \href{https://doi.org/10.1007/978-3-031-18576-2_8}{https://doi.org/10.1007/978-3-031-18576-2\underline{\space}8}.}}
\maketitle
\begin{abstract}
Unsupervised anomaly detection models which are trained solely by healthy data, have gained importance in the recent years, as the annotation of medical data is a tedious task. Autoencoders and generative adversarial networks are the standard anomaly detection methods that are utilized to learn the data distribution.
However, they fall short when it comes to inference and evaluation of the likelihood of test samples. We propose a novel combination of generative models and a probabilistic graphical model. After encoding image samples by autoencoders,
the distribution of data is modeled by Random and Tensorized Sum-Product Networks ensuring exact and efficient inference at test time.
We evaluate different autoencoder architectures in combination with Random and Tensorized Sum-Product Networks on mammography images using patch-wise processing and observe superior performance over utilizing the models standalone and state-of-the-art in anomaly detection for medical data.

\keywords{Anomaly Detection \and Generative Models \and Sum-Product Networks \and Mammography.}
\end{abstract}
\section{Introduction}
Acceleration of the detection and segmentation of anomalous tissue by automated computer aided approaches is a key for enhancing cancer screening programs. It is especially important for mammography screening, as breast cancer is the most common cancer type and the leading cause of death in women worldwide~\cite{wild2020}.
Training an artificial neural network in a supervised way needs a high amount of pixel-wise annotated data. As data annotation is very costly, methods which involve as less annotation as possible, are of high demand. Anomaly detection approaches are good representatives of this type, as they only utilize healthy cases for learning, and anomalous spots are detected as a deviation from the learned data distribution. The deviation is measured either by straight-forward metrics such as reconstruction error of input and output samples or by more sophisticated constructs such as log-likelihood in probabilistic models.

Unsupervised anomaly detection methods have been evaluated on a plethora of different pathologies and medical imaging modalities. A state-of-the-art method in this area is f-AnoGAN~\cite{schlegl2019f} which leverages Generative Adversarial Networks (GANs) to model an implicit distribution of healthy images and detect outliers via a custom anomaly score based on reconstruction performance. f-AnoGAN has been utilized to detect anomalies in OCT scans~\cite{schlegl2019f}, Chest X-rays~\cite{bhatt2021unsupervised} and 3D Brain scans~\cite{simarro2020unsupervised}. However, it requires the training of a separate encoder module to obtain latent codes of images, which are used by the generator for reconstruction. The autoencoder architecture (AE) on the other hand jointly trains an encoder and decoder and is thus able to directly map an input to its corresponding latent representation. AE variants have been applied to lesion detection in mammography images~\cite{wei2018anomaly} and brain scans~\cite{marimont2021anomaly,zimmerer2019unsupervised} as well as head~\cite{sato2018primitive} and abdomen~\cite{marimont2021anomaly} CT scans. However, the practical applicability of all those models is limited by the fact that the respective anomaly scores are not easily interpretable by a human decision maker. Here, to remedy the situation, it would be desirable for the model to provide some degree of certainty for its decision. To this end, density estimation models can be employed. Such models learn an explicit probability density function from the training data and assume that anomalous samples are located within low-density regions. Examples are the application of Gaussian Mixture Models (GMMs)~\cite{bowles2017brain} for brain lesion detection as well as Bayesian U-Nets for OCT anomaly detection~\cite{seebock2019exploiting}. Although these approaches are similar to ours, they are tailored to specific image modalities and can thus not be directly and easily applied to our domain.

In this work, we introduce a novel and general method for anomaly detection which combines AE architectures with probabilistic graphical models called Sum-Product Networks (SPNs).
A recent powerful SPN-type called Random and Tensorized SPN (RAT-SPN)~\cite{peharz2020} was chosen, as they are easy to integrate into deep learning frameworks and are trained by GPU-based optimization. More than that, while standard and variatonal AEs only provide either simple image reconstruction based or the evidence lower bound based inference metrics for anomaly detection, SPNs, on the other hand, allow \emph{exact} and \emph{efficient} likelihood inference by imposing special structural constraints on modeling the data distribution. Therefore, we compare different AE architectures combined with RAT-SPNs on the unsupervised lesion and calcification detection use-case in public mammography scans and demonstrate improved performance.

\begin{figure}
\centering
\begin{tikzpicture}[>=latex, scale=0.9, transform shape,	font=\scriptsize,
	across/.style={-,path picture={ 
 		\draw[black] (path picture bounding box.east) -- (path picture bounding box.west) (path picture bounding box.south) -- (path picture bounding box.north);
	}}, 
	pcross/.style={-,path picture={ 
  		\draw[black] (path picture bounding box.north east) -- (path picture bounding box.south west) (path picture bounding box.south east) -- (path picture bounding box.north west);
	}},
	sumnode/.style={circle, across, draw, fill=white, text width=\twidth, minimum size=\twidth},
	prodnode/.style={circle, pcross, draw, fill=white, text width=\twidth,minimum size=\twidth},
	leafnode/.style={circle, draw, minimum size=\twidth, inner sep=0pt, align=center, fill=white},
	->]
	
	\newcommand{\mammopatches}[5]{
		\node[draw=#3!50, fill=#3!5, thick, minimum height=3.6*#1, minimum width=1.3*#1] (#5box) at (#2,1.1*#1) {};
		\node (#5img1) at (#2,0) {\includegraphics[width=#1]{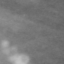}};
		\node (#5img2) at (#2,1.1*#1) {\includegraphics[width=#1]{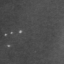}};
		\node (#5img3) at (#2,2.2*#1) {\includegraphics[width=#1]{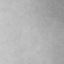}};
		\node (#5label) at (#2,-1) {#4};
	}
	\def\pdf{\usebox{\genericfilt}};
	
	\setlength{\figsize}{0.05\textwidth}
	\mammopatches{\figsize}{0}{color14}{Input Patches}{in};
	\node[draw=color13!50, fill=color13!5, minimum width=15pt, minimum height=15pt] (latent) at (3,\figsize+1) {$\mathbf{Z}_n$};
	\mammopatches{\figsize}{6}{color6}{Reconstructions}{out};
	
	\draw (inbox) -- (0.75,\figsize+1);
	\draw (2.25,\figsize+1) -- (latent);
	\draw (latent) -- (3.75,\figsize+1);
	\draw (5.25,\figsize+1) -- (outbox);
	
	\node[minimum width=0.5cm, minimum height=1.5cm] (X1) at (0,\figsize+1) {};
	\draw[fill=color9!5, draw=color9!50] ([xshift=0.5cm]X1.north east)coordinate(aux) -- ([xshift=2cm,yshift=0.2cm]X1.east)coordinate(aux1) -- ([xshift=2cm,yshift=-0.2cm]X1.east) -- ([xshift=0.5cm]X1.south east) -- cycle; 
	\node[minimum width=0.5cm, minimum height=1.5cm] (X2) at (3,\figsize+1) {};
	\draw[fill=color15!5, draw=color15!50] ([xshift=2cm]X2.north east)coordinate(aux) -- ([xshift=0.5cm,yshift=0.2cm]X2.east)coordinate(aux1) -- ([xshift=0.5cm,yshift=-0.2cm]X2.east) -- ([xshift=2cm]X2.south east) -- cycle; 

	\node (encoder) at (1.5,\figsize+1) {Encoder};	
	\node (decoder) at (4.5,\figsize+1) {Decoder};

	\node[leafnode] (leaf1) at (8.5,25+\figsize) {\pdf};
	\node[] at (8.5,15+\figsize) {\vdots};
	\node[leafnode] (leaf2) at (8.5,\figsize) {\pdf};
	\node[draw=black!50, inner sep=3pt, fit={(leaf1) (leaf2)}] (box1) {};
	\node[rotate=270] (box1label) at (8,\figsize+12.5) {$\{Z_1,Z_3\}$};
	
	\node[leafnode] (leaf3) at (8.5,\figsize-20) {\pdf};
	\node[]  at (8.5,\figsize-30) {\vdots};
	\node[leafnode] (leaf4) at (8.5,\figsize-45) {\pdf};
	\node[draw=black!50, inner sep=3pt, fit={(leaf3) (leaf4)}] (box2) {};
	\node[rotate=270] (box2label) at (8,\figsize-32.5) {$\{Z_0,Z_2\}$};
	
	\node[prodnode] (prod1) at (9.5,25+\figsize) {};
	\node[prodnode] (prod2) at (9.5,10+\figsize) {};
	\node[] (proddots) at (9.5,\figsize-1) {\vdots};
	\node[prodnode] (prod3) at (9.5,15-\figsize) {};
	\node[prodnode] (prod4) at (9.5,-\figsize) {};
	\node[draw=black!50, inner sep=3pt, fit={(prod1) (prod4)}] (box3) {};
	
	\foreach \x in {1,...,4}
		\draw (prod\x) -- (box1);
	
	\foreach \x in {1,...,4}
		\draw (prod\x) -- (box2);

	\node[sumnode, label=right:{$p(E(\mathbf{x}))$}] (root) at (10.5, \figsize-5) {};
	
	\foreach \x in {1,...,4}
		\draw (root) -- (prod\x);
	\draw (root) --  (9.7,\figsize-5);
	\node (hiddens) at (11.7, \figsize-5) {};
	
	\node[sumnode, label=right:{sum node}] (sum) at (0,3.5) {};
	\node[prodnode, label=right:{product node}] (prod) at (2.5,3.5) {};
	\node[leafnode, label=right:{leaf node}] (leaf) at (5,3.5) {\pdf};
	\node[label=right:{latent variable}] (z) at (7.5, 3.5) {$\mathbf{Z}_n$};
	\node (hiddenz) at (9.6,3.5) {};
	
	\begin{scope}[on background layer]
		\node[draw=black!50, dashed, rounded corners=.25cm, inner sep=5pt, fit={(encoder) (decoder) (inimg3) (inlabel) (outimg3) (outlabel)}, label={\small Autoencoder} ] {};
		\node[draw=black!50, dashed, rounded corners=.25cm, inner sep=5pt, fit={(hiddens) (box1) (box2) (box1label) (box2label)}, label={\small Sum-Product Network} ] (spn) {};
		\node[draw=black!50, dashed, rounded corners=.25cm, inner xsep=10pt, inner ysep=5pt, fit={(sum) (hiddenz)} ] {};
	 \end{scope};
	 
	 \draw (latent) --  (3,-1.5) -- (7.35,-1.5) -- (7.35, \figsize-10) -- (spn);
	 \node at (5.3,-1.75) {$\{Z_0, Z_1, Z_2, Z_3\}$};
	
\end{tikzpicture}
\caption{The encoder of an AE outputs a low-dimensional latent representation $\mathbf{z}_n$ of healthy mammography patches. This representation serves then as input to a SPN that learns the corresponding probability distribution $p(E(\mathbf{x}))$. The likelihood of test samples is predicted over the same pipeline using trained models.
}
\label{fig:system_architecture}
\end{figure}
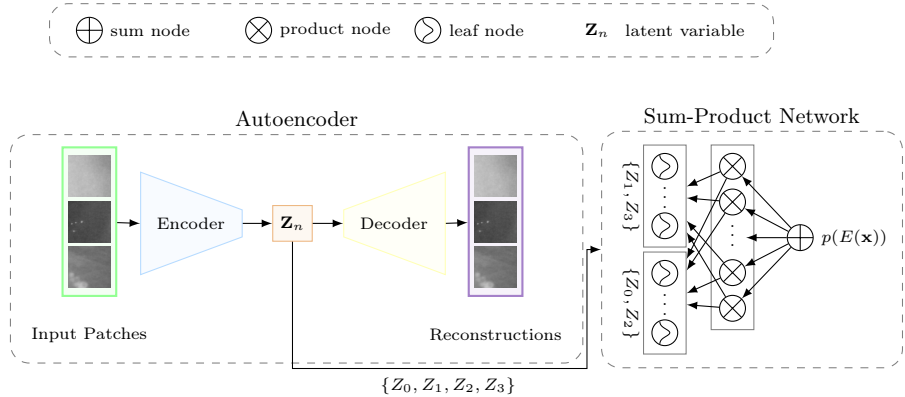

\section{Methods}
Our approach learns the healthy data distribution in a patch-wise fashion. First, the dimensionality of patch data is reduced by an AE and the likelihood for membership to the data distribution is approximated by a RAT-SPN. During inference, the learned model is applied on test images patch by patch, where at every position the likelihood yields the anomaly score. As the models capture the distribution of healthy data, this score should be significantly different at anomalous image positions. We compare the performance of our pipeline to either the reconstruction or variational inference based approximation quality of the AE models. The different AE architectures are described in Section~\ref{sec:ae} and the RAT-SPNs are covered in Section~\ref{sec:spn}. A system overview is provided in Figure~\ref{fig:system_architecture}.
\subsection{Autoencoders}
\label{sec:ae}
\paragraph{Convolutional AEs (CAEs)~\cite{masci2011}} utilize convolutional blocks to map high dimensional image data $\mathbf{x} \in \mathbb{R}^{N\times N}$ into a lower dimensional latent space $\mathbf{z} \in \mathbb{R}^{M}$ using an encoder by $\mathbf{z} = E(\mathbf{x})$ and reconstruct it utilizing a decoder model by $\hat{\mathbf{x}} = D(E(\mathbf{x}))$. The compression and reconstruction process is learned by minimizing the reconstruction loss $\mathcal{L}_\text{CAE} = \ell_2(\mathbf{x},\hat{\mathbf{x}})$ where $\ell_2$ signalizes the mean squared error (MSE). Computation of $\mathcal{L}_\text{CAE}$ for test samples yields the anomaly score at inference.
\paragraph{Variational Autoencoders (VAEs) \cite{kingma2013}} are equipped by the same building blocks as CAEs when applied to images, but additionally, they aim to approximate the true posterior distribution $p(\mathbf{z}|\mathbf{x})$ in the encoder $E$ by a simpler and more tractable distribution $q(\mathbf{z}|\mathbf{x})$. This is achieved by minimizing the KL divergence $D_{KL}(q(\mathbf{z}|\mathbf{x}) \left|\right| p(\mathbf{z}|\mathbf{x}) )$ between the two distributions. The decoder $D$, on the other hand, learns the posterior $p(\mathbf{x}|\mathbf{z})$ and reconstructs $\mathbf{x}$ from a given $\mathbf{z}$ by maximizing the log-likelihood $\log p(\mathbf{x}|\mathbf{z})$.  
The overall objective to minimize is called the evidence lower bound (ELBO) and it can be formulated as follows:
\begin{equation}
\mathcal{L}_{VAE} = \mathbb{E}_{q(\mathbf{z}|\mathbf{x})}\left[\log p(\mathbf{x}|\mathbf{z})\right] - \beta D_{KL}(q(\mathbf{z}|\mathbf{x}) \left|\right| p(\mathbf{z}) ).
\end{equation}

The MSE was utilized as the reconstruction loss for $\log p(\mathbf{x}|\mathbf{z})$ and $\beta$ was set to $0.1$ according to~\cite{higgins2017} which is a weighting factor between the two terms. Thus, our models are called $\beta$VAEs~\cite{higgins2017}. $\mathcal{L}_{VAE}$ is utilized as the anomaly score for a given test sample during inference.

\paragraph{Vector Quantised-Variational Autoencoder (VQVAE)~\cite{van2017neural}} is a VAE variant which differs from the original in a crucial aspect: it uses discrete instead of continuous variables to represent the latent space. Discretization is realised by mapping the encoder output $E(\mathbf{x})$ to the index of the closest vector $e_i$ in the latent embedding space $\mathbf{e} \in \mathbb{R}^{K \times B}$, where $K$ is the number of distinct discrete values and $B$ is the dimension of each embedding vector $e_i$. The posterior variational distribution $q(\mathbf{z}|\mathbf{x})$ is one-hot-encoded in such a way that $q(\mathbf{z} = k|\mathbf{x}) = 1$, with $k = \argmin_i||E(\mathbf{x}) - e_i||_2$. The mapping of $E(\mathbf{x})$ to the nearest embedding vector $e_i$ is defined as $E_q(\mathbf{x}) = e_k$ with $k = \argmin_i||E(\mathbf{x}) - e_i||_2$. The loss formulation consists of three parts, each aiming to optimize a different aspect of the model:
\begin{equation}
\mathcal{L}_\text{VQVAE} = \log p(\mathbf{x}|E_q(\mathbf{x})) + ||sg[E(\mathbf{x})] - \mathbf{e}||_2^2 + \lambda ||E(\mathbf{x}) - sg[\mathbf{e}]||_2^2.
\end{equation}

The first term is the reconstruction loss, for which MSE was again chosen. The remaining terms are concerned with learning an optimal embedding space. The codebook loss, the second term, attempts to move the embedding vectors closer to the encoder output, whereas the third term, the commitment loss, attempts the inverse and forces the encoder output to be closer to the closest embedding vector $\mathbf{e}$. $sg[\cdot]$ is the stopgradient operator and prevents its operand from being updated during back-propagation. $\lambda$ is a weighting factor for the commitment loss, which we set to 0.25, following \cite{van2017neural}. The anomaly score for a given test sample is determined by calculating its reconstruction loss.

\subsection{Sum-Product Networks}
\label{sec:spn}
Sum-product networks (SPNs)~\cite{poon2011sum} are tractable probabilistic models of the family of probabilistic circuits~\cite{choi2020probabilistic} and allow various probabilistic queries to be computed efficiently and exactly.
For consistency with recent works, we will introduce SPNs based on the formalism in \cite{trapp2019bayesian}.
An SPN on a set of RVs $\mathbf{Z} = \{Z_j\}^D_{j=1}$ is a tuple $(\G, \psi)$ consisting of a computational graph $\G$, which is a directed acyclic graph, and a scope function $\psi$ mapping from the set of nodes in $\G$ to the set of all subsets of $\mathbf{Z}$ including $\mathbf{Z}$.
The computational graph of an SPN is typically composed of sum nodes, product nodes, and leaf nodes.
Sum nodes compute a weighted sum of their children, i.e., $(\Sum(\vz) = \sum_{\Node \in \ch(\Sum)} \theta_{\Sum,\Node}\, \Node(\vz) )$, product nodes compute a product of their children, i.e., $(\Prod(\vz) = \prod_{\Node \in \ch(\Prod)} \Node(\vz) )$, and leaf nodes are tractable multivariate or univariate probability distributions or indicator functions.
The scope function assigns each node a scope (subset of $\mathbf{Z}$ or $\mathbf{Z}$) and ensures that the SPN fulfils certain structural properties, guaranteeing that specific probabilistic queries can be answered tractably.
In this work we will focus on SPNs that are \emph{smooth} and \emph{decomposable}, we refer to \cite{choi2020probabilistic} for a detailed discussion.
Moreover, we consider a representation of the SPN in form of a random and tensorized region-graph called RAT-SPNs~\cite{peharz2020} and employ the implementation based on Einstein summation as proposed in \cite{peharz2020einsum}.
The region-graph is parametrized by the number of root nodes $C$, input distributions $I$ as well as the graph depth $D$ and the number of parallel SPN instances, or recursive splits, $R$. By choosing this parameters, RAT-SPNs with arbitrary complexity may be constructed. From a given region-graph it is possible to obtain the underlying SPN structure in terms of its computational graph and scope function exactly, for more details see \cite{trapp2019bayesian,peharz2020}. A simple region-graph with $C=1$, $I=2$, $D=1$ and $R=1$ and the underlying SPN is illustrated in Figure \ref{fig:system_architecture}.
In a generative learning setting, like ours, the optimal network parameters $w$ are found by applying (stochastic) Expectation Maximization (EM) to maximize the log-likelihood $LL$ of the training samples:
\begin{equation}
\mathbf \\LL{(w)} = \frac{1}{N} \sum_{n=1}^{N}\log{\Sum(\vz_n)}. 
\end{equation} 
The likelihood obtained by evaluating a trained RAT-SPN model on a given test sample is used as our anomaly score.	

\section{Experimental Setup}

\subsection{Datasets}
\label{sec:datasets}
We train our models on the Digital Database for Screening Mammography (DDSM) \cite{ddsm}, a collection of 2620 mammography exams, with each exam consisting of multiple images. The images in this dataset are categorized according to the type of diagnosis, either into \emph{healthy} or into a \emph{cancer} type (i.e. malignant, benign). As we want to learn a healthy model, we selected only the 695 healthy exams containing 2798 images for training purposes. From each image, 120 patches of 64$\times$64 pixels (px), the resolution also used by~\cite{schlegl2019f}, were extracted; half of these containing internal breast tissue, the other half were sampled along the breast contour. We evaluated all methods against a selection of cancerous mammograms from the Curated Breast Imaging Subset of DDSM (CBIS-DDSM) \cite{cbisddsm} which provides improved annotations of masses and calcifications for images from DDSM. 79 scans were selected from the mass and 30 scans from the calcification test set with the restriction that the annotation mask area must be smaller than 4-times of our patch-area for masses and it should contain the whole calcifications. The healthy training images consisted of equally distributed dense and non-dense tissues, whereas the mass test cases had a ratio of $14\%/86\%$ and the calcification test samples a ratio of $40\%/60\%$ (dense/non-dense).

\subsection{Training}
The 2798 healthy images were split into 90\% training and 10\% validation images (with no patient overlap) for training both the generative (CAE, $\beta$VAE, VQVAE) and the RAT-SPN models. Following~\cite{masci2011}, all of our generative models had an architecture with 32-64-128 2D convolutional layers with 5$\times$5 kernels and a stride of 2 in the encoder and 2D transposed convolutional layers in the decoder. The VQVAE model had additional 6 residual blocks with 128 number of filters and the dimensionality of the embedding vector was 64. All models had 64 units for the latent representation and were trained with a batch size of 64. CAEs and $\beta$VAEs were trained for 100 epochs with a learning rate of 1e-5, whereas VQVAEs converged to an optimum after 20 epochs with a learning rate of 1e-4. 
The best-fit RAT-SPN parameters of $D=1, R=50, I=45$ were found utilizing the 10\% validation images, possible values were taken from the supplement of~\cite{peharz2020}.
The RAT-SPN setup was the same for all generative models and it was trained by the EM-algorithm for 50 epochs with batch size of 64 and learning rate of 1e-4.

\begin{figure}
\centering
\includegraphics[width=1.0\textwidth]{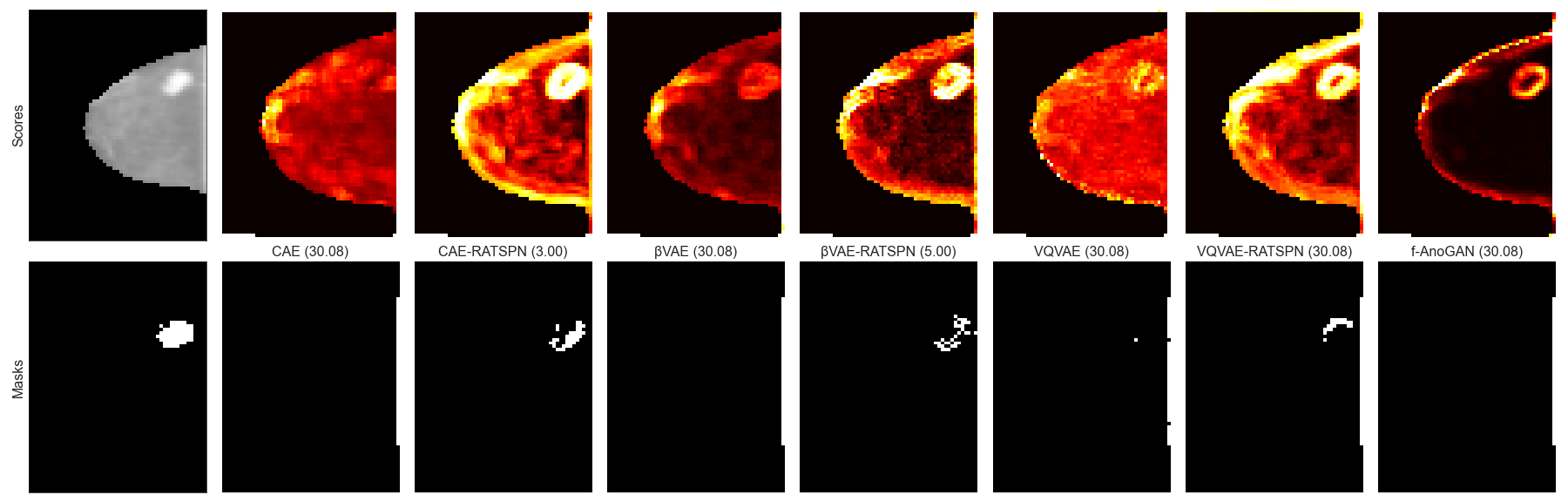}
\caption{Anomaly detection results. First column shows the mammography scan (top) and anomaly ground-truth (bottom). The remaining columns depict an anomaly score heatmap (top) and segmentation mask (bottom) for each model. The respective Hausdorff distance to the ground-truth (px) is displayed after each method name in brackets.}
\label{fig:anomaly_segmentations_mass_test}
\end{figure}

\subsection{Evaluation}
We evaluated our methods on the 79 mass and the 30 calcification test images. Anomaly score assignment was performed in a lower dimensional image space than the original resolution, and thus, patches were sampled around every $16^{th}$ pixel per image. Only breast tissue pixels were considered using pre-segmentations of the breast-area in every image.
We measured the Area Under the ROC-Curve (AUC) of the computed anomaly scores in two ways, either considering all pixels from all test images at once (pixel-wise) or doing it for each image separately and calculating the average over all test samples additionally (image-wise). In order to measure the capability of the methods for segmentation of the anomalous regions, we apply the Hausdorff distance (H) image-wise to assess the pixel distance between masks generated by our models and the provided CBIS-DDSM ground-truth. It measures the maximum of the distances from any annotated point in one mask to the nearest point in the other mask, thus the smaller it is, the closer the match between prediction and ground-truth.

\section{Results and Conclusion}
We compare the anomaly detection performance of three AE models in their standalone configuration as well as with a RAT-SPN extension. Our AE models cover continuous (CAE and $\beta$VAE) and discrete (VQVAE) feature spaces. Additionally, we trained and evaluated an f-AnoGAN model (state-of-the-art) in its default configuration. The results are illustrated in Figure~\ref{fig:anomaly_segmentations_mass_test} and in Table~\ref{table:results_lesions}. 

\begin{table}
\begin{center}
\begin{tabular}{|c|c|c|c|c|}
\hline
\multirow{2}{*}{\textit{Test Data}} & \multirow{2}{*}{\textit{Model}} & \multicolumn{1}{c|}{\textbf{Pixel-wise}} & \multicolumn{2}{c|}{\textbf{Image-wise}} \\
\cline{3-5}
                                &  & \textit{AUC} & \textit{AUC} & \textit{H} \\
\hline
\hline
\multirow{7}{*}{\textbf{Masses}} & CAE & 0.53 & 0.58$\pm$0.25 &  30.80$\pm$12.57 \\

																			  & CAE-RATSPN & \textbf{0.88} & \textbf{0.88$\pm$0.10} & \textbf{30.10$\pm$14.50} \\ 
																				\cline{2-5}
																
																		    & $\beta$VAE & 0.80 & 0.83$\pm$0.14 & 30.78$\pm$12.79 \\
																				
																				& $\beta$VAE-RATSPN & 0.88 & 0.88$\pm$0.11 & 29.07$\pm$14.57  \\
																				\cline{2-5}
																						
																				& VQVAE & 0.67 & 0.67$\pm$0.19 &  33.37$\pm$12.53 \\
																						
																				& VQVAE-RATSPN & \textbf{0.82} & \textbf{0.84$\pm$0.13} & \textbf{32.21$\pm$14.01} \\
																				\cline{2-5}
																						& f-AnoGAN & 0.86 & 0.85$\pm$0.12 & 30.41$\pm$11.91  \\
\hline
\hline
\multirow{7}{*}{\textbf{Calcifications}} 		& CAE & 0.65 & 0.77$\pm$0.21 & 32.79$\pm$10.82  \\

																		        & CAE-RATSPN & 0.72 & 0.78$\pm$0.16 & 33.03$\pm$13.74  \\
																						\cline{2-5}
											
																	        	& $\beta$VAE & 0.73 & 0.80$\pm$0.17 & 29.73$\pm$12.56  \\
																						
																						& $\beta$VAE-RATSPN & 0.66 & 0.73$\pm$0.17 &  30.72$\pm$13.07 \\
																						\cline{2-5}
																						
																						& VQVAE & 0.69 & 0.79$\pm$0.17 & 31.79$\pm$12.66  \\
																						
																						& VQVAE-RATSPN & 0.68 & 0.75$\pm$0.19 & 33.34$\pm$10.61  \\
																						\cline{2-5}
																						& f-AnoGAN & 0.67 & 0.74$\pm$0.20 & 34.95$\pm$6.98  \\
\hline																
\end{tabular}
\end{center}
\caption{Anomaly detection results utilizing different models. Metrics are computed either over pixels or images. Next to AUC scores average Hausdorff (H) distances (px) between anomaly segmentations and ground-truth masks were computed. Segmentations are calculated after score thresholding by $99^{th}$-percentile. Statistically significantly better performance (based one image-wise AUCs) between standalone and RAT-SPN extended models are depicted in bold (p<0.01).}
\label{table:results_lesions}
\end{table}

\begin{figure}
\centering
\begin{subfigure}{.24\textwidth}
  \centering
  \includegraphics[width=\textwidth]{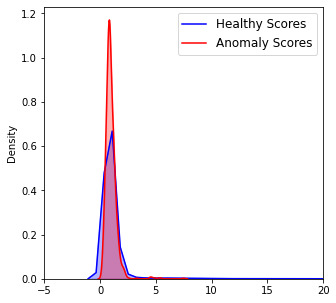}
	\caption{}
\end{subfigure}%
\begin{subfigure}{.245\textwidth}
  \centering
  \includegraphics[width=\textwidth]{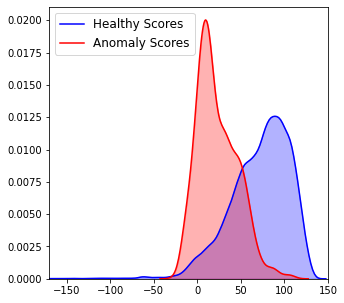}
	\caption{}
\end{subfigure}
\begin{subfigure}{.23\textwidth}
  \centering
  \includegraphics[width=\textwidth]{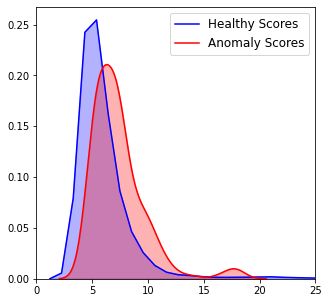}
	\caption{}
\end{subfigure}%
\begin{subfigure}{.24\textwidth}
  \centering
  \includegraphics[width=\textwidth]{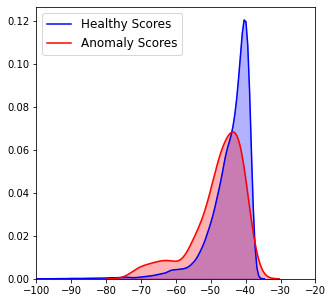}
	\caption{}
\end{subfigure}
\caption{Distribution of healthy and anomalous scores on the masses (a,b) and calcifications datasets (c,d) for CAE without (a) and with RAT-SPN extension (b), for $\beta$VAE without (c) and with RAT-SPN extension (d).}
\label{fig:score_dist_comparison_cae-spn_mass-calc_test}
\end{figure}

For the mass test set, the best performing model was the $\beta$VAE-RATSPN with $0.88$ pixel-wise and average image-wise AUCs, and an average $H$-distance of $29.07$~px. Statistically significant superior performances over standalone models were achieved by CAE-RATSPNs and VQVAE-RATSPNs. Except for VQVAE-RATSPN, all RAT-SPN extended models performed better than f-AnoGAN in terms of image-wise AUC, although there were no statistically significant differences (cf. Table~\ref{table:results_lesions}).
It is also visible in Table~\ref{table:results_lesions}, that RAT-SPNs applied to continuous features yielded better results than the discrete version. Furthermore, it is depicted in Figure~\ref{fig:score_dist_comparison_cae-spn_mass-calc_test} a) and b), that attaching RAT-SPN models to generative models facilitate a better discrimination between healthy and anomalous tissue by increasing the gap between their respective distributions.

Moreover, standalone $\beta$VAEs were the best performing models for the calcification test set with an $0.73$ pixel-wise and $0.80$ average image-wise AUC, and an average $H$-distance of $29.73$~px. It is in general visible, that all models reflect a consistently poorer performance for this data. It is due to the fact that this set contains a higher proportion of dense breasts than the mass collection (see Sec.~\ref{sec:datasets}) and most of the small calcifications were generally hard to detect accurately by all models in images dominated by dense tissue.
On the other hand, the standalone versions performed here better than the ones with RAT-SPN extension except for the CAE setup, but no statistically significant differences were discovered based on the image-wise AUC scores (cf. Table~\ref{table:results_lesions}). This behavior is well visualized by the score distribution plots of the best performing standalone $\beta$VAE and $\beta$VAE-RATSPN versions in Figure~\ref{fig:score_dist_comparison_cae-spn_mass-calc_test} c) and d). All models except for $\beta$VAE-RATSPN yielded better image-wise AUCs than f-AnoGAN, although no statistically significant differences were detected (cf. Table~\ref{table:results_lesions}).

In summary, we have introduced a novel unsupervised anomaly detection method that extends various AE architectures with a RAT-SPN module. This approach is a promising avenue for generating exact likelihoods and incorporating them into the detection of different anomalies, such as masses and calcifications in mammography scans. Our experiments suggest that our method works remarkably well on mass cases, while it struggles when applied to calcification samples. We interpret that this is due to the presence of a larger proportion of dense tissue in the latter dataset. In future work we want to analyze, how this problem can be eliminated and in particular whether increasing the input resolution has a positive effect on the performance.

\subsubsection{Acknowledgements} 
VRVis is funded by BMK, BMDW, Styria, SFG, Tyrol and Vienna Business Agency in the scope of COMET - Competence Centers for Excellent Technologies (879730) which is managed by FFG. Thanks go to AGFA HealthCare, project partner of VRVis, for providing valuable input. \mbox{Martin Trapp} acknowledges funding from the Academy of Finland (347279).

\bibliographystyle{splncs04}
\bibliography{refs}

\end{document}